\documentclass{article} 
\usepackage{iclr2024_conference,times}

\usepackage{amsmath,amsfonts,bm}

\def\eqref#1{equation~\ref{#1}}

\def\1{\bm{1}}

\DeclareMathAlphabet{\mathsfit}{\encodingdefault}{\sfdefault}{m}{sl}
\SetMathAlphabet{\mathsfit}{bold}{\encodingdefault}{\sfdefault}{bx}{n}

\usepackage{hyperref}
\usepackage{url}
\usepackage{booktabs}
\usepackage{graphicx}
\usepackage{xspace}
\usepackage{comment}

\title{AfriHG: News headline generation for \\ African Languages}

\author{Toyib Ogunremi~\textsuperscript{1}, Serah Akojenu~\textsuperscript{1}, Anthony Soronnadi~\textsuperscript{1}, Olubayo Adekanmbi~\textsuperscript{1}, \\
\textbf{David Ifeoluwa Adelani}~\textsuperscript{2}\\
\textsuperscript{1} Data Scientists Network  \\
\textsuperscript{2} University College London \\
\texttt{toyib@datasciencenigeria.ai}, \ \ \texttt{d.adelani@ucl.ac.uk} \\
}

\newcommand*{\yoruba}{Yor\`ub\'a\xspace}

\iclrfinalcopy 
\begin{document}

\maketitle

\begin{abstract}
This paper introduces AfriHG---a news headline generation dataset created by combining from XLSum and MasakhaNEWS datasets focusing on 16 languages widely spoken by Africa. We experimented with two seq2eq models (mT5-base and AfriTeVa V2), and Aya-101 LLM. Our results show that Africa-centric seq2seq models such as AfriTeVa V2 outperform the massively multilingual mT5-base model. Finally, we show that the performance of fine-tuning AfriTeVa V2 with 313M parameters is competitive to prompting Aya-101 LLM with more than 13B parameters.

\end{abstract}

\section{Introduction}

Summarization is an established text generation task in NLP, this involves extracting the most important parts or sections that can describe the important or general section of the whole document.
There are two popular types of text summarization~\citep{Vipul_6754792}: (1) extractive summarization and  (2) abstractive summarization. \textbf{Extractive} summarization focues on concatenating extracts or excerpts of a document as summary while abstractive summarization focuses on \textit{paraphrasing} or reframing and information compression which capture the main key points in the document. \textbf{Abstractive} summarization is popular these days because of its ability to generate diverse and coherent summary of the document being summarized with the application of Neural network architectures~\citep{shukla23}. In this paper, we focus on \textbf{news headline generation}---which can be regarded as  a special type of abstractive summarization as it involves carving out information that can serve as the headline of the news or a \textit{one-line summary}. We created \textsc{AfriHG}--- a new dataset with article-headline pair obtained from popular news articles like BBC, VOA, and Isolezwe covering 16 widely spoken languages in Africa. We evaluated the performance of this news headline generation task on two multilingual T5 models: mT5~\citep{xue-etal-2021-mt5} and AfriTeVa V2~\citep{oladipo-etal-2023-better}. Our evaluation shows a very good performance similar to text summarization, and  AfriTeVa V2 performing better than mT5 since it has been pre-trained on many more African languages. However, languages with non-Latin script have a very low ($<4.0$ Rouge~\citep{lin-2004-rouge} score 1) on news headline generation unlike the news summarization task where they achieved very good results ($>20.0$ Rouge score 1). Lastly, we evaluated Aya LLM~\citep{ustün2024aya}---an instruction-tuned mT5-XXL model on several tasks including  summarization and headline generation of XL-SUM~\citep{hasan-etal-2021-xl}. Our findings shows that AfriTeVa V2 with 313M is competitive to Aya with 13B parameters. For reproducibility, we release the code and data on GitHub.\footnote{\url{https://github.com/dadelani/AfriHG}}

\section{ Data and Languages covered}

\label{gen_inst}

This paper utilizes news articles and their respective headlines as published by \textbf{BBC News}, a division of the British Broadcasting Corporation (BBC). Founded on November 14, 1922, BBC News has evolved into the world's largest broadcast news organization, renowned for its comprehensive global coverage and journalistic integrity. It offers a blend of radio, internet, and television broadcasts, making it an ideal source for diverse and high-quality news content. Notably, BBC News extends its reach beyond high-resource languages like English, French and Arabic, offering content in 10 African languages namely: Amharic, Hausa, Igbo, Kirundi, Nigerian-Pidgin, Oromo, Somali, Kiswahili, Tigrinya, and \yoruba. Other news sources are Isolezwe for isiXhosa and isiZulu language, and VOA for Shona language.

\paragraph{Benchmark data} We collected summarization data from XL-Sum \citep{hasan-etal-2021-xl}---a dataset containing 1 million news article-summary pairs in 44 languages scraped from BBC news, is the first publicly available abstractive summarization dataset. This dataset also included the headline information for each article. We merged this data with the corpus used for creating MasakhaNEWS corpus \citep{adelani2023masakhanews}---a news topic classification dataset covering 16 languages widely spoken in Africa. 
We merged these two datasets together and removed duplicates. \textit{We retained the development and test split of the original XL-SUM dataset}  for a fair comparison between the tasks of news summarization and headline generation while also maintaining the 80\%-10\%-10\% TRAIN-DEV-TEST split for the languages not covered by BBC/XL-SUM i.e. chiShona, isiXhosa, and isiZulu. After combining the datasets, we filtered very short articles that are less than 10 characters.

\paragraph{AfriHG dataset}
We refer to our new dataset as \textbf{AfriHG}---it covers 16 languages spoken in Africa, 13 are from the BBC websites, including 10 African languages, Arabic, English, and French. The rest of the languages covered are from the MasakhaNEWS corpus: Shona (obtained from Voice of America)~\footnote{\url{https://www.voashona.com/}}, isiXhosa and isiZulu. The last two are from the Isolezwe website~\footnote{\url{https://www.isolezwelesixhosa.co.za/} and \url{https://www.isolezwe.co.za/}}.

\section{Baseline models}
We fine-tune two T5~\citep{2020t5} sequence-sequence models on the news headline generation tasks: mT5-base~\cite{xue-etal-2021-mt5} and AfriTeVa V2-base~\citep{oladipo-etal-2023-better}. We fine-tune the models using a batch size of 4, number of epochs 3, and the default learning rate of $5e-5$. All models are fine-tuned on Nvidia A10 single GPU using the HuggingFace framework~\citep{wolf-etal-2020-transformers}. We provide detailed information about the T5-models below: 

\paragraph{mt5-base}
mT5-base is a multilingual T5 model with 580M parameters. It was pretrained on the mC4 corpus with 6.63 trillion tokens covering 101 languages which covers 17 of languages spoken in Africa: Afrikaans, Amharic, Arabic, Chichewa, English, French, Hausa, Igbo, Malagasy, Shona, Sinhala, Somali, Sotho, Kiswahili, isiXhosa, \yoruba, isiZulu. 

\paragraph{Afriteva V2}
Afriteva V2 base is an extension of AfriTeVa model\citep{jude-ogundepo-etal-2022-afriteva} from 10 African languages to 20 widely spoken languages in Africa.
The base model contains about 313M parameters. Similar to mT5-base, it has 12 layers, 12 attention heads and 512 token sequence length. It was pre-trained on WURA dataset~\footnote{\url{https://huggingface.co/datasets/castorini/wura}}---a cleaned version of mC4 corpus with 29.3GB of data. 
AfriTeVa V2 covers all the languages in the AfriHG dataset.

\paragraph{Aya}
Aya is an instruction-tuned massively multilingual model~\citep{ustün2024aya} available in 101 languages. It is a large language model with 13B parameters built by fine-tuning an XXL-T5 model finetuned on Aya Dataset~\citep{singh2024aya} consisting of 25 African languages. Aya covers 12 of the languages evaluated in the AfriHG dataset.

\begin{table}[t]
\centering
\resizebox{\columnwidth}{!}{%
\begin{tabular}{lcr|r|rrr}
 & &  & \textbf{summary} & \multicolumn{3}{c}{\textbf{headline generation}}\\
 & &  \textbf{Data size} & \textbf{mT5-base} & \textbf{mT5-base} & \textbf{AfriTeVa V2 (base)} & \textbf{Aya}\\
\textbf{Language} & \textbf{script}&  \textbf{train / dev / test} & \textbf{R1 / R2 / RL} & \textbf{R1 / R2 / RL} & \textbf{R1 / R2 / RL} & \textbf{R1 / R2 / RL} \\ 
\midrule
Amharic (amh) & Ge'ez & 16,915 / 719 / 719 & 20.0 / 7.4 / 18.1 & 17.9 / 6.0 / 16.9  & 19.4 / 6.8 / 18.6 & \textbf{22.6 / 8.5 / 21.4}\\
Arabic (arb) & Arabic & 37,519 / 4,689 / 4689 & 35.0 / 14.8 / 29.2 & 25.3 / 9.0 / 23.2  & 24.1 / 8.2 / 24.1 & \textbf{28.1 / 10.8 / 25.4}\\
English (eng) & Latin & 311,694 / 11,535 / 11,535 & 37.6 / 15.2 / 29.9 & 36.1 / 13.5 / 31.9  & \textbf{37.9 / 14.5 / 33.4} & 32.3 / 11.7 / 28.2\\
French (fra) & Latin & 15,377 / 1,086 / 1,086 & 35.3 / 16.2 / 28.2 & 30.6 / 13.7/ 27.2  & 33.8 / 15.7 / 29.8 & \textbf{35.9 / 17.5 / 31.0}\\
Hausa (hau) & Latin & 17,174/ 802 / 802 & 39.4 / 17.7 / 31.7 & 30.2 / 11.1 / 26.9 & \textbf{35.0/ 14.9/ 31.5} & 34.5 / 14.8 / 30.1\\
Igbo (ibo) & Latin & 8,877 / 522 / 522 & 31.6 / 10.2 / 24.5 & 28.5 / 11.2 / 24.6 & 31.0 / 12.7/ 27.2 & \textbf{33.4 / 14.9 / 28.7}\\
Oromo (orm) & Latin & 16,417 / 757 / 757 & 18.7 / 6.2 / 16.2 & 15.7 / 4.7 / 14.8 & 18.8 / 6.5 / 17.6 & \textbf{19.6 / 6.7 / 18.3}\\
Pidgin (pcm) & Latin & 18,214 / 1,151 / 1,151 & 39.0 / 15.1 / 29.9 & 31.5 / 10.9 / 27.0 & \textbf{33.8 / 12.2 / 29.1} & 30.5 / 10.5 / 26.2\\
Kirundi (run) & Latin & 9,930 / 718 / 718 & 32.0 / 14.4 / 25.8 & 25.2 / 8.9 / 22.5 & \textbf{29.2 / 11.0 / 25.7} & 27.9 / 10.5 / 25.1\\
Shona (sna) & Latin & 9,573 / 1,064 / 1,064 &  - / - / - & 22.6 / 8.1 / 22.1 & \textbf{25.5 / 9.6 / 24.7} & 23.5 / 8.2 / 22.7\\
Somali (som) & Latin & 10,508 / 745 / 745 & 31.6 / 11.6 / 24.2 & 24.1 / 7.0 / 21.2  & 28.2 / 9.6 / 24.7 & \textbf{28.6 / 10.5 / 24.9}\\
Swahili (swa) & Latin & 18,914 / 987 / 987 & 37.7 / 17.9 / 30.9 &   33.0 / 13.4 / 29.1 & 37.6 / 15.8 / 33.1 & \textbf{38.9 / 16.7 / 33.9} \\
Tigrinya (tir) & Ge'ez & 12,351 / 681 / 681 & 25.3 / 8.0 / 21.2 & 21.9/ 7.0 / 19.8 & 26.3 / 8.7/ 23.7 & \textbf{25.7 / 8.6 / 22.8} \\
Xhosa (xho) & Latin & 10,440 / 1,305 / 1,305 & - / - / - &  13.0 / 4.0/ 12.7 & 15.2 /5.5 / 14.9 & \textbf{16.1 / 5.3 / 15.2} \\
Yoruba (yor) & Latin & 15,172 / 793 / 793 & 31.7 / 11.7 / 25.1 & 40.0 /14.8 / 31.4  & \textbf{42.0 / 16.2 / 33.1} & 36.0 / 13.5 / 28.3\\
Zulu (zul) & Latin & 14,209 / 1,776 / 1,776 & - / - / - & 16.0 / 5.0/ 15.5 & \textbf{17.8 /5.5 / 17.4} & 17.0 / 4.9 / 16.2\\
\midrule
Average & -- & - & - / - / - & 25.7 / 9.3 / 22.9 & \textbf{28.5 / 10.84 / 25.5} & 28.2 / 10.85 / 24.9\\
\bottomrule
\end{tabular}
}
\vspace{-3mm}
\caption{\textbf{Rouge score (R1/R2/RL) comparing news summarization and news headline generation}. The baseline summarization model results were obtained from \citep{hasan-etal-2021-xl}. 
- / - / - indicates the evaluation values that are not available in the XL-Sum}
\label{lab:lang_results}
\end{table}

\section{Results}
\autoref{lab:lang_results} compares the result of news summarization and headline generation. Comparing the two T5 models on headline generation, in all languages we evaluated on, AfriTeVa V2-base was consistently better than the mT5-base with average performance gain in Rouge score of $+2.8 / +1.5 / +2.6 $ (R1/R2/RL). This shows that AfriTeVa V2 is a better pre-trained language model for African languages since it has seen more during pre-training. For non-Latin scripts such as Arabic, Amharic and Tigrinya, we obtained low \texttt{rouge} scores, when we used the default \textit{auto tokenizer} provided by HuggingFace. However, switching to the language/script dependent tokenizer obtained a more realistic scores.\footnote{\url{https://huggingface.co/UBC-NLP/AraT5-base}} \footnote{\url{https://huggingface.co/rasyosef/bert-amharic-tokenizer}} \footnote{\url{https://huggingface.co/abrhaleitela/TigXLNet}} The gap in performance without the script specific tokenizer can be as high as $-20$ points difference in R1/RL performance. We leave this for further investigation. 


Finally, we evaluated on Aya, and show that it is competitive to AfriTeVa V2-base, achieving the best result for 9 out of the 16 languages. However, it still struggles with non-Latin script. While Aya LLM has a more than 43 times more parameters than AfriTeVa V2, and has seen the same training data, the performance are very similar. This shows that there is still some advantage in using the \textit{fine-tune paradigm} rather than \textit{prompting} when there is abundance of training data in this LLM age. 


\section{Conclusion}
In this paper, we presented AfriHG, an extended African news headline generation corpus compiled from XL-SUM dataset and MasakhaNEWS corpus. 
We added the MasakhaNEWS corpus to widen the language coverage while also providing more news-categories which are not available in the XL-Sum dataset for a more-inclusive evaluation. We performed experiments on this dataset using state-of-the-art multilingual pre-trained language models and demonstrated the capability of these seq2seq models in generating headlines for news articles in various African languages. Future work will focus on evaluating the performance of other LLMs such as GPT-4, LlaMa and Gemma model types.

\bibliography{iclr24_conference}

\begin{thebibliography}{12}
\providecommand{\natexlab}[1]{#1}
\providecommand{\url}[1]{\texttt{#1}}
\expandafter\ifx\csname urlstyle\endcsname\relax
  \providecommand{\doi}[1]{doi: #1}\else
  \providecommand{\doi}{doi: \begingroup \urlstyle{rm}\Url}\fi

\bibitem[Adelani et~al.(2023)Adelani, Masiak, Azime, Alabi, Tonja, Mwase, Ogundepo, Dossou, Oladipo, Nixdorf, Emezue, sana~al azzawi, Sibanda, David, Ndolela, Mukiibi, Ajayi, Moteu, Odhiambo, Owodunni, Obiefuna, Mohamed, Muhammad, Ababu, Salahudeen, Yigezu, Gwadabe, Abdulmumin, Taye, Awoyomi, Shode, Adelani, Abdulganiyu, Omotayo, Adeeko, Afolabi, Aremu, Samuel, Siro, Kimotho, Ogbu, Mbonu, Chukwuneke, Fanijo, Ojo, Awosan, Kebede, Sakayo, Nyatsine, Sidume, Yousuf, Oduwole, Tshinu, Kimanuka, Diko, Nxakama, Nigusse, Johar, Mohamed, Hassan, Mehamed, Ngabire, Jules, Ssenkungu, and Stenetorp]{adelani2023masakhanews}
David~Ifeoluwa Adelani, Marek Masiak, Israel~Abebe Azime, Jesujoba Alabi, Atnafu~Lambebo Tonja, Christine Mwase, Odunayo Ogundepo, Bonaventure F.~P. Dossou, Akintunde Oladipo, Doreen Nixdorf, Chris~Chinenye Emezue, sana~al azzawi, Blessing Sibanda, Davis David, Lolwethu Ndolela, Jonathan Mukiibi, Tunde Ajayi, Tatiana Moteu, Brian Odhiambo, Abraham Owodunni, Nnaemeka Obiefuna, Muhidin Mohamed, Shamsuddeen~Hassan Muhammad, Teshome~Mulugeta Ababu, Saheed~Abdullahi Salahudeen, Mesay~Gemeda Yigezu, Tajuddeen Gwadabe, Idris Abdulmumin, Mahlet Taye, Oluwabusayo Awoyomi, Iyanuoluwa Shode, Tolulope Adelani, Habiba Abdulganiyu, Abdul-Hakeem Omotayo, Adetola Adeeko, Abeeb Afolabi, Anuoluwapo Aremu, Olanrewaju Samuel, Clemencia Siro, Wangari Kimotho, Onyekachi Ogbu, Chinedu Mbonu, Chiamaka Chukwuneke, Samuel Fanijo, Jessica Ojo, Oyinkansola Awosan, Tadesse Kebede, Toadoum~Sari Sakayo, Pamela Nyatsine, Freedmore Sidume, Oreen Yousuf, Mardiyyah Oduwole, Tshinu Tshinu, Ussen Kimanuka, Thina Diko, Siyanda Nxakama, Sinodos
  Nigusse, Abdulmejid Johar, Shafie Mohamed, Fuad~Mire Hassan, Moges~Ahmed Mehamed, Evrard Ngabire, Jules Jules, Ivan Ssenkungu, and Pontus Stenetorp.
\newblock Masakhanews: News topic classification for african languages, 2023.

\bibitem[Dalal \& Malik(2013)Dalal and Malik]{Vipul_6754792}
Vipul Dalal and Latesh Malik.
\newblock A survey of extractive and abstractive text summarization techniques.
\newblock In \emph{2013 6th International Conference on Emerging Trends in Engineering and Technology}, pp.\  109--110, 2013.
\newblock \doi{10.1109/ICETET.2013.31}.

\bibitem[Hasan et~al.(2021)Hasan, Bhattacharjee, Islam, Mubasshir, Li, Kang, Rahman, and Shahriyar]{hasan-etal-2021-xl}
Tahmid Hasan, Abhik Bhattacharjee, Md.~Saiful Islam, Kazi Mubasshir, Yuan-Fang Li, Yong-Bin Kang, M.~Sohel Rahman, and Rifat Shahriyar.
\newblock {XL}-sum: Large-scale multilingual abstractive summarization for 44 languages.
\newblock In \emph{Findings of the Association for Computational Linguistics: ACL-IJCNLP 2021}, pp.\  4693--4703, Online, August 2021. Association for Computational Linguistics.
\newblock URL \url{https://aclanthology.org/2021.findings-acl.413}.

\bibitem[Lin(2004)]{lin-2004-rouge}
Chin-Yew Lin.
\newblock {ROUGE}: A package for automatic evaluation of summaries.
\newblock In \emph{Text Summarization Branches Out}, pp.\  74--81, Barcelona, Spain, July 2004. Association for Computational Linguistics.
\newblock URL \url{https://www.aclweb.org/anthology/W04-1013}.

\bibitem[Ogundepo et~al.(2022)Ogundepo, Oladipo, Adeyemi, Ogueji, and Lin]{jude-ogundepo-etal-2022-afriteva}
Odunayo~Jude Ogundepo, Akintunde Oladipo, Mofetoluwa Adeyemi, Kelechi Ogueji, and Jimmy Lin.
\newblock {A}fri{T}e{VA}: Extending ?small data? pretraining approaches to sequence-to-sequence models.
\newblock In \emph{Proceedings of the Third Workshop on Deep Learning for Low-Resource Natural Language Processing}, pp.\  126--135, Hybrid, July 2022. Association for Computational Linguistics.
\newblock \doi{10.18653/v1/2022.deeplo-1.14}.
\newblock URL \url{https://aclanthology.org/2022.deeplo-1.14}.

\bibitem[Oladipo et~al.(2023)Oladipo, Adeyemi, Ahia, Owodunni, Ogundepo, Adelani, and Lin]{oladipo-etal-2023-better}
Akintunde Oladipo, Mofetoluwa Adeyemi, Orevaoghene Ahia, Abraham Owodunni, Odunayo Ogundepo, David Adelani, and Jimmy Lin.
\newblock Better quality pre-training data and t5 models for {A}frican languages.
\newblock In Houda Bouamor, Juan Pino, and Kalika Bali (eds.), \emph{Proceedings of the 2023 Conference on Empirical Methods in Natural Language Processing}, pp.\  158--168, Singapore, December 2023. Association for Computational Linguistics.
\newblock \doi{10.18653/v1/2023.emnlp-main.11}.
\newblock URL \url{https://aclanthology.org/2023.emnlp-main.11}.

\bibitem[Raffel et~al.(2020)Raffel, Shazeer, Roberts, Lee, Narang, Matena, Zhou, Li, and Liu]{2020t5}
Colin Raffel, Noam Shazeer, Adam Roberts, Katherine Lee, Sharan Narang, Michael Matena, Yanqi Zhou, Wei Li, and Peter~J. Liu.
\newblock Exploring the limits of transfer learning with a unified text-to-text transformer.
\newblock \emph{Journal of Machine Learning Research}, 21\penalty0 (140):\penalty0 1--67, 2020.
\newblock URL \url{http://jmlr.org/papers/v21/20-074.html}.

\bibitem[Shukla et~al.(2023)Shukla, Barange, Shahabade, Pandey, and Bavkar]{shukla23}
Karishma Shukla, Kartik Barange, Prajakta Shahabade, Akanksha Pandey, and Dnyaneshwar Bavkar.
\newblock Abstractive text summarization using transformer based approach.
\newblock 06 2023.
\newblock \doi{10.55041/IJSREM22369}.

\bibitem[Singh et~al.(2024)Singh, Vargus, Dsouza, Karlsson, Mahendiran, Ko, Shandilya, Patel, Mataciunas, OMahony, Zhang, Hettiarachchi, Wilson, Machado, Moura, Krzemiński, Fadaei, Ergün, Okoh, Alaagib, Mudannayake, Alyafeai, Chien, Ruder, Guthikonda, Alghamdi, Gehrmann, Muennighoff, Bartolo, Kreutzer, Üstün, Fadaee, and Hooker]{singh2024aya}
Shivalika Singh, Freddie Vargus, Daniel Dsouza, Börje~F. Karlsson, Abinaya Mahendiran, Wei-Yin Ko, Herumb Shandilya, Jay Patel, Deividas Mataciunas, Laura OMahony, Mike Zhang, Ramith Hettiarachchi, Joseph Wilson, Marina Machado, Luisa~Souza Moura, Dominik Krzemiński, Hakimeh Fadaei, Irem Ergün, Ifeoma Okoh, Aisha Alaagib, Oshan Mudannayake, Zaid Alyafeai, Vu~Minh Chien, Sebastian Ruder, Surya Guthikonda, Emad~A. Alghamdi, Sebastian Gehrmann, Niklas Muennighoff, Max Bartolo, Julia Kreutzer, Ahmet Üstün, Marzieh Fadaee, and Sara Hooker.
\newblock Aya dataset: An open-access collection for multilingual instruction tuning, 2024.

\bibitem[Wolf et~al.(2020)Wolf, Debut, Sanh, Chaumond, Delangue, Moi, Cistac, Rault, Louf, Funtowicz, Davison, Shleifer, von Platen, Ma, Jernite, Plu, Xu, Scao, Gugger, Drame, Lhoest, and Rush]{wolf-etal-2020-transformers}
Thomas Wolf, Lysandre Debut, Victor Sanh, Julien Chaumond, Clement Delangue, Anthony Moi, Pierric Cistac, Tim Rault, Rémi Louf, Morgan Funtowicz, Joe Davison, Sam Shleifer, Patrick von Platen, Clara Ma, Yacine Jernite, Julien Plu, Canwen Xu, Teven~Le Scao, Sylvain Gugger, Mariama Drame, Quentin Lhoest, and Alexander~M. Rush.
\newblock Transformers: State-of-the-art natural language processing.
\newblock In \emph{Proceedings of the 2020 Conference on Empirical Methods in Natural Language Processing: System Demonstrations}, pp.\  38--45, Online, October 2020. Association for Computational Linguistics.
\newblock URL \url{https://www.aclweb.org/anthology/2020.emnlp-demos.6}.

\bibitem[Xue et~al.(2021)Xue, Constant, Roberts, Kale, Al-Rfou, Siddhant, Barua, and Raffel]{xue-etal-2021-mt5}
Linting Xue, Noah Constant, Adam Roberts, Mihir Kale, Rami Al-Rfou, Aditya Siddhant, Aditya Barua, and Colin Raffel.
\newblock m{T}5: A massively multilingual pre-trained text-to-text transformer.
\newblock In \emph{Proceedings of the 2021 Conference of the North American Chapter of the Association for Computational Linguistics: Human Language Technologies}, pp.\  483--498, Online, June 2021. Association for Computational Linguistics.
\newblock \doi{10.18653/v1/2021.naacl-main.41}.
\newblock URL \url{https://aclanthology.org/2021.naacl-main.41}.

\bibitem[Üstün et~al.(2024)Üstün, Aryabumi, Yong, Ko, D'souza, Onilude, Bhandari, Singh, Ooi, Kayid, Vargus, Blunsom, Longpre, Muennighoff, Fadaee, Kreutzer, and Hooker]{ustün2024aya}
Ahmet Üstün, Viraat Aryabumi, Zheng-Xin Yong, Wei-Yin Ko, Daniel D'souza, Gbemileke Onilude, Neel Bhandari, Shivalika Singh, Hui-Lee Ooi, Amr Kayid, Freddie Vargus, Phil Blunsom, Shayne Longpre, Niklas Muennighoff, Marzieh Fadaee, Julia Kreutzer, and Sara Hooker.
\newblock Aya model: An instruction finetuned open-access multilingual language model, 2024.

\end{thebibliography}
\bibliographystyle{iclr2024_conference}


\end{document}